\documentclass{article}

    \PassOptionsToPackage{numbers, compress}{natbib}
    \usepackage[final]{neurips_2022}

\usepackage[utf8]{inputenc} % allow utf-8 input
\usepackage[T1]{fontenc}    % use 8-bit T1 fonts
\usepackage{hyperref}       % hyperlinks
\usepackage{url}            % simple URL typesetting
\usepackage{booktabs}       % professional-quality tables
\usepackage{amsfonts}       % blackboard math symbols
\usepackage{nicefrac}       % compact symbols for 1/2, etc.
\usepackage{microtype}      % microtypography
\usepackage{xcolor}         % colors
\usepackage{xspace}

\usepackage{amsmath}
\usepackage{amsfonts}
\usepackage{amssymb}
\usepackage{wrapfig}
\usepackage{subcaption}
\usepackage{multirow}
 \usepackage{mathtools} 

\usepackage{verbatim}

\usepackage{anyfontsize}

\usepackage{microtype}
\usepackage{graphicx}
\usepackage{booktabs} % for professional tables
\usepackage{multirow}
\usepackage{amsmath,amssymb}
\usepackage{booktabs}
\usepackage{caption,subcaption}

\usepackage{xcolor}
\definecolor{mygreen}{HTML}{3cb44b}
\definecolor{skyblue}{HTML}{beffff}
\definecolor{lightgreen}{HTML}{90ee90}

\usepackage{color, colortbl}

\definecolor{emerald}{rgb}{0.31, 0.78, 0.37}

\usepackage{tcolorbox}
\usepackage{enumitem}
\setitemize{itemsep=10pt,topsep=0pt,parsep=0pt,partopsep=0pt}
\usepackage{colortbl}

\usepackage{xcolor}
\definecolor{mygreen}{HTML}{3cb44b}
\colorlet{myyellow}{green!10!orange!90!}
\makeatletter

\usepackage{tikz}
\usetikzlibrary{arrows,shapes,snakes,automata,backgrounds,fit,petri}
\usepackage{adjustbox}

\newcommand{\RN}[1]{%
	\textup{\lowercase\expandafter{\it \romannumeral#1}}%
}
\usepackage{tabu}
\newcommand{\beq}{\vspace{0mm}\begin{equation}}
\newcommand{\eeq}{\vspace{0mm}\end{equation}}
\newcommand{\beqs}{\vspace{0mm}\begin{eqnarray}}
\newcommand{\eeqs}{\vspace{0mm}\end{eqnarray}}
\newcommand{\barr}{\begin{array}}
\newcommand{\earr}{\end{array}}

\usepackage{color, colortbl}
\definecolor{Gray}{gray}{0.93}

\usepackage{lipsum}

\newcommand\blfootnote[1]{%
  \begingroup
  \renewcommand\thefootnote{}\footnote{#1}%
  \addtocounter{footnote}{-1}%
  \endgroup
}

\usepackage{pifont}% http://ctan.org/pkg/pifont

\usepackage{makecell}

\usepackage{xcolor,amsmath}
\usepackage[linesnumbered,ruled,vlined]{algorithm2e}
\DontPrintSemicolon

\usepackage{xcolor}
\definecolor{mygreen}{HTML}{3cb44b}

\SetKwComment{Comment}{\color{green!50!black}\# }{}

\SetKwProg{Function}{def}{:}{}

\SetKwProg{For}{for}{:}{}
\SetKwProg{If}{if}{:}{}
\newcommand{\VarSty}[1]{\textnormal{\ttfamily\color{blue!90!black}#1}\unskip}

\usepackage{listings} % for code block

\definecolor{Gray}{gray}{0.5}
\definecolor{LGray}{gray}{0.9}
\definecolor{darkblue}{RGB}{94,110,186}
\definecolor{darkGreen}{RGB}{92, 148, 110}
\definecolor{myblue}{RGB}{14, 121, 178}

\newcommand{\red}[1]{\textcolor{red}{#1}}

\newcommand{\darkGreen}[1]{\textcolor{darkGreen}{#1}}

\def\logo{\makebox[0pt][l]{\hspace{0pt}\raisebox{-0.3ex}{\includegraphics[height=19pt]{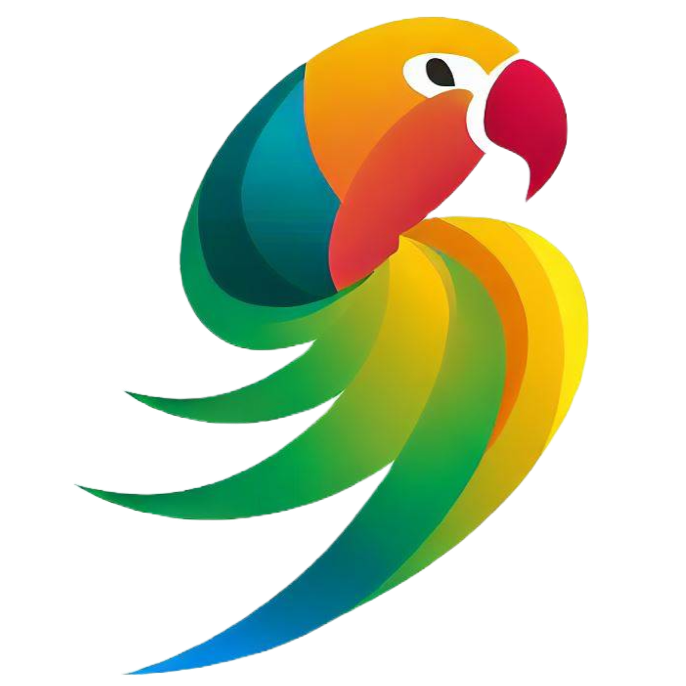}}}}
\newcommand{\modelname}{VideoChat}
\def\logoblue{\makebox[0pt][l]{\hspace{0pt}\raisebox{-0.3ex}{\includegraphics[height=10pt]{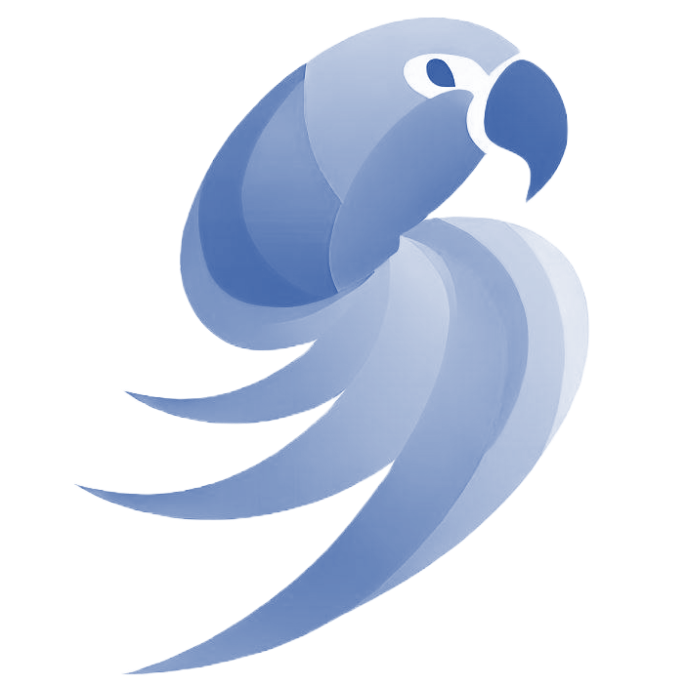}}}}
\def\logotext{\makebox[0pt][l]{\hspace{0pt}\raisebox{-0.3ex}{\includegraphics[height=10pt]{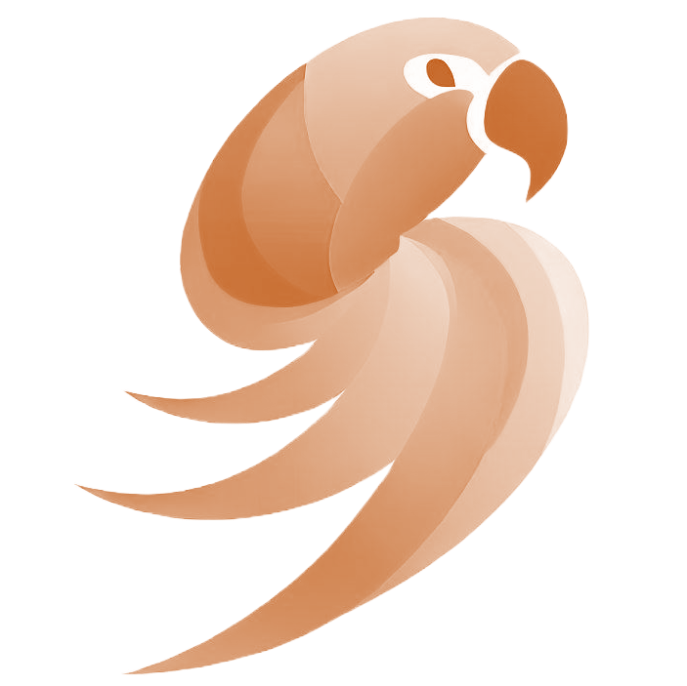}}}}
\def\logollava{\makebox[0pt][l]{\hspace{0pt}\raisebox{-0.3ex}{\includegraphics[height=10pt]{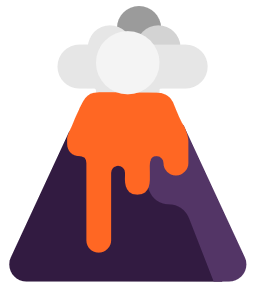}}}}
\def\logominigpt{\makebox[0pt][l]{\hspace{0pt}\raisebox{-0.3ex}{\includegraphics[height=10pt]{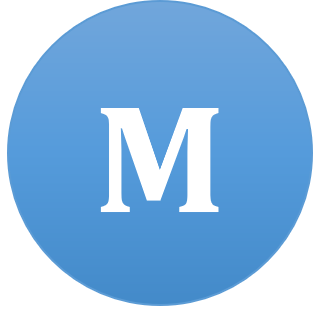}}}}
\def\logoowl{\makebox[0pt][l]{\hspace{0pt}\raisebox{-0.3ex}{\includegraphics[height=10pt]{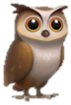}}}}

\title{\darkGreen{V}ideo\darkGreen{Chat}\logo\ \ \ \ : Chat-Centric Video Understanding}

\author{%
  \hspace{-0.25cm}\textbf{Kunchang Li$^{*1,4}$, Yinan He$^{*1}$, Yi Wang$^{*\dagger 1}$, Yizhuo Li$^{1,3}$, Wenhai Wang$^{1}$} \\ \hspace{-0.25cm}\textbf{Ping Luo$^{3,1}$, Yali Wang$^{\dagger 4,1}$, Limin Wang$^{\dagger 2,1}$, Yu Qiao$^{\dagger 1}$} \\
 \hspace{-0.25cm}$^1$OpenGVLab, Shanghai AI Laboratory \quad $^2$Nanjing University \quad $^3$The University of Hong Kong\\
 \hspace{-0.25cm}$^4$Shenzhen Institutes of Advanced Technology, Chinese Academy of Sciences \\
 \\
 {\url{https://github.com/OpenGVLab/Ask-Anything}} \\
}

\begin{document}
\maketitle

\begin{abstract}
In this paper, we initiate an attempt of developing an end-to-end chat-centric video understanding system, coined as VideoChat. It integrates video foundation models and large language models via a learnable neural interface, excelling in spatiotemporal reasoning, event localization, and causal relationship inference. To instructively tune this system, we build a video-centric instruction dataset, composed of thousands of videos associated with detailed descriptions and conversations. This dataset emphasizes spatiotemporal reasoning and captures causal relationships, providing a valuable asset for training our chat-centric video understanding system. Preliminary qualitative experiments demonstrate the potential of our system across a broad spectrum of video applications, which could serve as a simple prototype system for future research on chat-centric video understanding. 
% Access our code and data at \href{https://github.com/OpenGVLab/Ask-Anything}{https://github.com/OpenGVLab/Ask-Anything}.
\end{abstract}

\section{Introduction}
Videos offer a remarkably close representation of how humans consistently perceive the visual world. Intelligent video understanding is crucial for various real-world applications, such as human-robot interaction, autonomous driving, and intelligent surveillance. However, current paradigms in video understanding are limited by task-specific tuning of pre-trained video foundation models, restricting a general spatiotemporal comprehension for video content. \blfootnote{* Equal contribution.\ \ \ \ $\dagger$ Corresponding authors.}

Vision-centric multimodal dialogue systems have recently emerged as an essential research area~\cite{kosmos}. By utilizing a pre-trained large language model (LLM), an image encoder, and additional learnable modules, these systems can deeply understand images (e.g., recognizing memes or jokes) and perform image-related tasks through multi-round dialogues with user queries~\cite{llava,minigpt,mplugowl}. This revolutionizes numerous visual applications, but existing systems have yet to formally address video-centric tasks from a data-centric perspective using learning machines.

Our initial video-centric multimodal dialogue system \footnote{\href{https://github.com/OpenGVLab/Ask-Anything}{\textcolor{red}{https://github.com/OpenGVLab/Ask-Anything, released on April 15, 2023.}}} %%记得补上日期
formulates video understanding as a natural language processing (NLP) question-answering, 
by textualizing video content with open-sourced visual models.
%vision classification/detection/caption models. 
Despite demonstrating decent performance in short-term scenarios with clear objects and actions, transforming videos into textual descriptions inevitably results in visual information loss and over-simplification of spatiotemporal complexities. Additionally, almost all utilized vision models struggle with spatiotemporal reasoning, event localization, and causal relationship inference within videos.

To tackle these challenges, we improve our initial dialogue system and introduce a groundbreaking chat-centric video understanding system that leverages state-of-the-art techniques from both video and language domains. Our approach creates a full loop, integrating video and language foundation models in a learnable manner from a model perspective, and provides all techniques required to learn the system from a data perspective. 

We begin by presenting our novel video-centric multimodal dialogue system. 
We propose an innovative system architecture that combines video foundation models and large language models (LLMs) through a learnable neural interface. By a two-stage lightweight training (with only spatiotemporal and video-language alignment modules) on large-scale video-text datasets and self-built video instruction ones, our method excels in spatiotemporal perception \& reasoning, and causal inference, marking the first attempt to create a fully learnable and efficient video understanding system that facilitates effective communication.

We introduce a novel video-centric multimodal instruction fine-tuning dataset. We create a unique dataset comprising thousands of videos paired with detailed textual descriptions and conversations generated using dense captions fed to ChatGPT in temporal order. This dataset emphasizes spatiotemporal objects, actions, events, and causal relationships, offering a valuable resource for training video-centric multimodal dialogue systems.

Through these contributions, our work pioneers new frontiers in video and natural language processing integration. By developing a new and effective chat-centric video understanding dialogue system, we pave the way for a wide range of applications across various domains while setting a standard for future research in this field. Our research not only pushes the boundaries of video understanding and reasoning but also offers protocols for both academic and industrial communities.
\section{Related Work}

\paragraph{Video Foundation Models}
Large-scale video-text pretraining coupled with downstream task fine-tuning has emerged as the standard paradigm in the video-language domain ~\cite{MiechASLSZ20,cpd, merlot,videoclip,umt,uniformerv2,merlot,videoclip,hu2022scaling,dou2022empirical,shen2021much,yao2021filip,videobert,actbert,wang2022internvideo,chen2022internvideo}. Early methods~\cite{videobert,actbert} employed pretrained visual and language encoders to derive offline video and text features; however, more recent approaches~\cite{cpd,MiechASLSZ20,hu2022scaling,dou2022empirical,videomae,wang2023videomae} have demonstrated the effectiveness of end-to-end training. Additionally, prevalent techniques often encompass two or three pretraining tasks, such as masked language modeling~\cite{lavender}, video-text matching~\cite{allinone}, video-text contrastive learning~\cite{videoclip,wang2022internvideo}, masked video modeling~\cite{videomae,wang2023videomae,wang2022internvideo} and video-text masked modeling~\cite{violet}. Within the realm of video multimodal tasks, VIOLET~\cite{violet} integrates masked language and masked video modeling, while All-in-one~\cite{allinone} suggests a unified video-language pretraining methodology using a shared backbone, and LAVENDER~\cite{lavender} consolidates the tasks through masked language modeling. Although these approaches yield impressive results in multimodal benchmarks, their training relies on limited video-text data, which leads to difficulties in video-only tasks such as action recognition. On the other hand, MERLOT Reserve~\cite{zellers2022merlot} compiles 20 million video-text-audio pairs for training joint video representations via contrastive span matching, thereby establishing state-of-the-art outcomes in video recognition and visual commonsense reasoning.

\paragraph{Large Language Models}
Recent advances in large language models (LLMs)~\cite{openai2022chatgpt,openai2023gpt4,brown2020gpt3,touvron2023llama,chiang2023vicuna,stablelm,alpaca} have showcased remarkable capabilities such as language generation, in-context learning, etc. These abilities enable LLMs to tackle complex tasks with user prompts in a zero-shot fashion. GPT-3~\cite{brown2020gpt3} shows notable zero-shot performance across numerous benchmarks. InstructGPT models~\cite{InstructGPT} are finetuned using datasets containing prompts with corresponding human-annotated desired behavior. This results in better alignment with users, improved output quality compared to GPT-3, increased truthfulness, and reduced risks. Instruction-tuned models also present remarkable generalization capacity for zero-shot tasks. Therefore, instruction-tuning~\cite{mishra2021cross,chung2022scaling} is crucial in leveraging LLMs' potential. Besides of GPT family~\cite{openai2022chatgpt,openai2023gpt4,brown2020gpt3}, there are multiple LLMs, including OPT~\cite{opt}, LLaMA~\cite{touvron2023llama}, MOSS~\cite{moss}, and GLM~\cite{zeng2022glm}, providing high-performance, open-source resources that can be finetuned for various purposes. For instance, Alpaca~\cite{alpaca} proposes a self-instruct framework to instruction-tune LLaMA models without heavily relying on human-authored instruction data.

\paragraph{LLMs for Multimodal Understanding}
The accomplishments of LLMs have accelerated the creation of AI systems that merge vision models with LLMs to enable multimodal reasoning and action~\cite{llava,minigpt,mplugowl,openflamingo,li2022blip,internchat,wu2023visual,Yang2023MMREACTPC,shen2023hugginggpt,liang2023taskmatrix,internchat}. Flamingo \cite{openflamingo} pioneered this approach by capitalizing on both vision and language models using web-scale image-text interwoven data, unveiling exceptional zero-shot image-text abilities in a conversational format for the first time. The study in \cite{kosmos} demonstrates that Kosmos-1 models are naturally equipped to tackle a broad array of perception-intensive tasks, including visual dialogue, visual explanation, visual question answering, image captioning, basic math equations, OCR, and zero-shot image classification using descriptions. Visual instruction tuning introduces an innovative technique for refining large language models on visual instruction tasks, enabling pretrained BLIP and Vicuna to nearly match GPT-4 level conversation performance for image-based tasks~\cite{llava}. MiniGPT-4 is a multimodal large language model, fine-tuned on multimodal tasks, and exhibits respectable zero-shot image comprehension in dialogues~\cite{minigpt}.
\section{\modelname}

\begin{figure*}[t]
    \centering
    \includegraphics[width=1\textwidth]{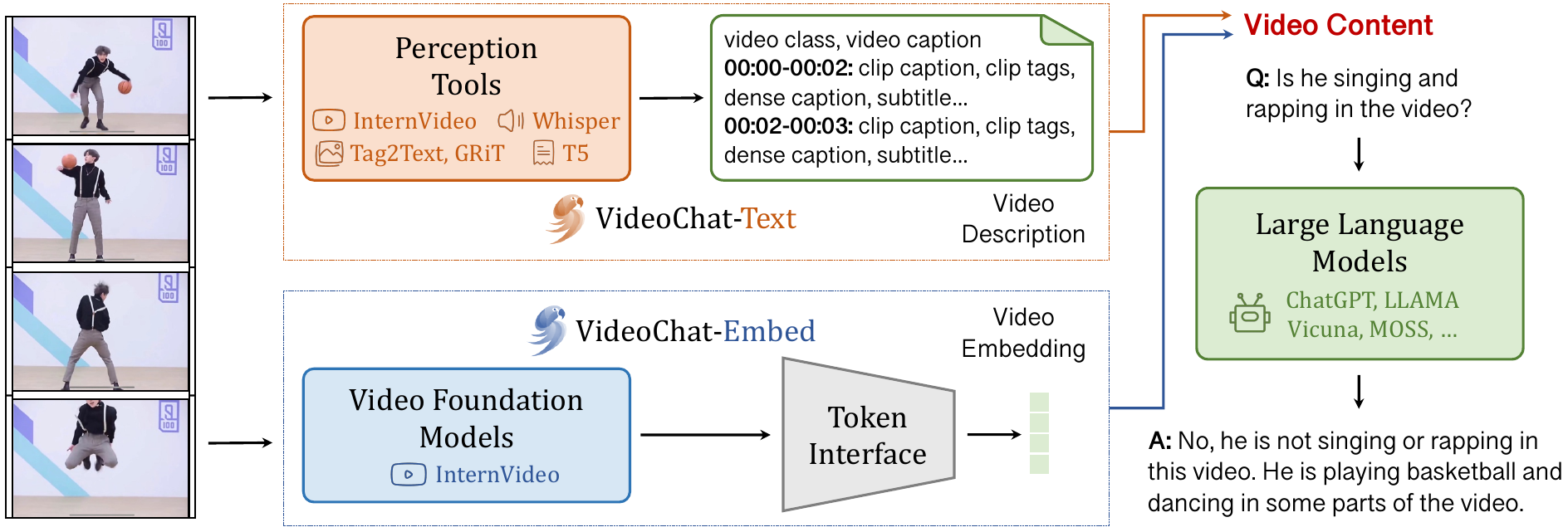}
    % \vspace{-0.3cm}
    \caption{
    \textbf{Framework.} 
    \textbf{VideoChat-Text} textualizes videos in stream. \textbf{VideoChat-Embed} encodes videos as embeddings.
    Both video content can be input in LLMs for multimodal understanding.
    }
    \label{fig:framework}
    \vspace{-0.3cm}
\end{figure*}

{\modelname} unifies video-related tasks into the formulation of multiple-round video question answering, 
\textcolor{red}{in which tasks are defined by words in a live inference and no or a few instances are given for learning.}
%这句话是个啥意思？ no or 这是啥？
In this formulation, we treat an LLM as a universal video task decoder, turning video-related descriptions or embeddings into human-understandable text. This procedure is user-friendly in employing foundation models to address various video applications.

Formally, we extract concepts from videos using vision models as:
\begin{equation}
[\mathbf{E}]_{i}^{j} = f_{\text{img}}^{j}(\mathbf{I}_{i}) \quad \text{or} \quad \mathbf{E}^{j} = f_{\text{vid}}^{j}(\mathbf{V}) \quad \text{w.r.t.} \quad \mathbf{V}=[\mathbf{I}_{i}]_{i=1,2,...,T},
\end{equation}
where $\mathbf{E}$ denotes a text description or embedding according to context, $f_{\text{img}}^{j}$ denotes the $j_\text{th}$ image model to predict human-readable annotations or visual feature, while $\mathbf{I}$ and $\mathbf{V}$ denote an image and video, respectively. Then we decode the task prediction from a LLM based on user's question as:
\begin{equation}
    \mathbf{W}_t^a = f_{\text{llm}}(\mathbf{E}|\mathbf{W}_{\leq t}^{q},\mathbf{W}_{<t}^{a}),
\end{equation}
where $\mathbf{W}_t^a$ and $\mathbf{W}_{\leq t}^{q}$ stand for the answers from the LLM at the round $t$ and all questions given by users before round $t$, respectively. $f_{\text{llm}}$ denotes an LLM model.

In technical terms, an ideal end-to-end chat-centric video understanding system should utilize a video/vision  model (an encoder) to convert visual sequences into latent features for LLM, guaranteeing the system's overall differentiability. Prior to this, we verify the efficacy of LLM as a universal video task interpreter through our proposed {\modelname}-Text (Section \ref{sec:videochat_text}). This method transforms videos into textual streams for subsequent discrimination/reasoning tasks using LLMs by incorporating various open-source vision models. While {\modelname}-Text can tackle typical spatiotemporal tasks such as spatial and temporal perception, it falls short in comprehending intricate temporal reasoning and causal inference. Therefore, we introduce {\modelname}-Embed (Section \ref{sec:videochat_embed}), a multimodal system that combines both video and language foundation models. Finetuned with video instruction data, it significantly enhances performance in higher-order temporal assignments. We will describe these two approaches in the following sections.

\begin{table*}[t]\centering
\begin{minipage}{0.99\columnwidth}\vspace{0mm}    
    \centering
    \begin{tcolorbox} 
        \centering
        \hspace{-6mm}
        \begin{tabular}{p{0.99\columnwidth}}
        \hspace{1mm}
        \begin{minipage}{0.99\columnwidth}
        Video Class, Video Caption \\
        \red{\texttt{00:00-00:02}}\ \  Clip Caption, Clip Tag, Dense Caption, Video Subtitle... \\
        \red{\texttt{00:02-00:03}}\ \  Clip Caption, Clip Tag, Dense Caption, Video Subtitle... \\
        \red{\texttt{00:03-00:06}}\ \  Clip Caption, Clip Tag, Dense Caption, Video Subtitle... \\
        $...$ \\ \rule[0.25\baselineskip]{\textwidth}{1pt}
        \begin{center}\includegraphics[width=0.5\linewidth]{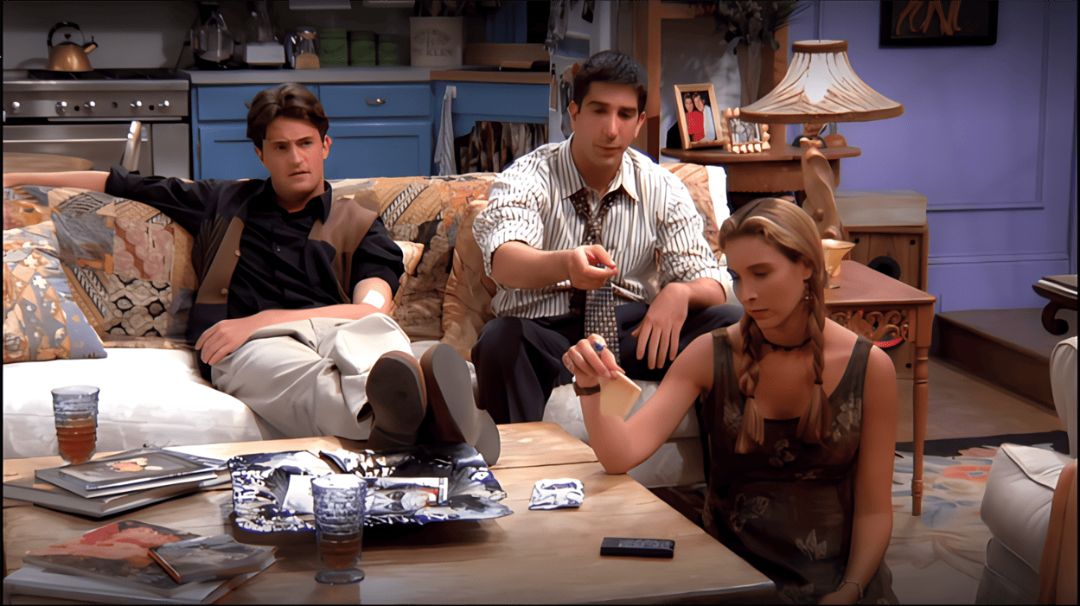}\end{center}
        answering questions, a man and a woman sitting on a couch in a living room with a table in front of them.\\
        00:00-00:11 a man and a girl sitting on a couch in a living room. \\a lamp with a white shadea woman sitting at a table: [446, 155, 710, 476]; man wearing a plaid shirt: [361, 44, 581, 337]; man sitting on couch: [10, 63, 324, 350]; the tie is grey: [441, 150, 486, 280]; a glass of beer: [38, 305, 77, 367]; a stack of magazines: [28, 350, 180, 394]; a white tablecloth: [0, 334, 626, 476]; stainless steel oven: [1, 55, 150, 142]; a brown tie on a man: [144, 168, 191, 270]; the couch is white: [0, 119, 730, 472]; a gray binder: [0, 377, 157, 411]; a white couch: [768, 350, 848, 477]; a lamp with a white shade: [582, 26, 713, 195];\\
        00:00-00:02: Hey, Pheebs, you gonna have the rest of that Pop-Tart?\\
        00:02-00:03: Pheebs?\\
        00:03-00:09: Does anyone want the rest of this Pop-Tart?\\
        00:09-00:11: Hey, I might.
        \end{minipage}
        \end{tabular}
    \end{tcolorbox}
    \vspace{-2mm}
    \caption{\textbf{Video description with perception models.} We use perception models to obtain the video clip action class, clip/video caption, and dense caption as well as a subtitle. Then we organize them in a template to generate textualizing videos.}
    \label{tab:video_description}
    \vspace{-2mm}
\end{minipage}
\end{table*}

\subsection{\modelname-Text\logotext\ \ \ \  : {\modelname} by Textualizing Videos in Stream}
\label{sec:videochat_text}

We employ several vision models to convert video data into textual format. Subsequently, we create purpose-built prompts to temporally structure the predicted text. Ultimately, we rely on a pretrained LLM to address user-specified tasks by responding to questions based on video text descriptions.

In particular, for a given video, we use $\mathtt{ffmpeg}$ to extract key frames from the video at a low $\mathtt{FPS}$, resulting in $T$ video frames and associated audio. By feeding the extracted frames and audio into various models, we acquire action labels, frame summaries, video tags, comprehensive descriptions, object positional coordinates, video narratives, timestamps, and other segment-related details. We then consolidate related content in the captions considering the timing and generate a timestamped video text description. We will first outline the vision models and prompt schematics employed, and then conclude with an analysis of {\modelname}-Text.

\subsubsection{Perception Models}
Utilizing a combination of video and image models~\cite{wang2022internvideo,raffel2020t5,uniformerv2,grit,tag2text,whisper}, we analyze videos from various aspects such as actions~\cite{wang2022internvideo,uniformerv2,umt} (with InternVideo \cite{wang2022internvideo}), objects~\cite{wang2022internimage,tag2text}, object annotations with positions~\cite{grit}, and more. While the majority of these models' outputs are comparatively independent, we utilize the pretrained T5 language model~\cite{raffel2020t5} to refine their descriptions for improved clarity. Moreover, we integrate the Whisper~\cite{whisper} speech recognition model into {\modelname}-Text to capitalize on audio data within videos, further enhancing the richness of video descriptions.

\subsubsection{Prompt System}

We process the video into different visual models to obtain different textualizing  videos and then organize them together in a template (Table \ref{tab:video_description}) as inputs to an LLM.
Then, we present the LLM with the context that we instruct it to pretend to watch the given video through the input formatted texts (the structured video knowledge from perception models) and then chat with users. 
Such prompt is shown in Table \ref{tab:system_prompt}.

\begin{table*}[h!]\centering
\begin{minipage}{0.99\columnwidth}\vspace{0mm}    
    \centering
    \begin{tcolorbox} 
        \centering
        % \small
        \hspace{-6mm}
        \begin{tabular}{p{0.99\columnwidth}}
        \hspace{1mm}
        \begin{minipage}{0.99\columnwidth}
        You are a chatbot that conducts conversations based on video contexts. You mainly answer based on the given contexts, and you can also modify the content according to the tag information, and you can also answer the relevant knowledge of the person or object contained in the video. \textbf{The timing description is a description every \VarSty{$1/FPS$} second, so that you can convert it into time. When describing, please mainly refer to the timing description. Dense caption is to give content every five seconds, you can disambiguate them in timing.} But you don't create a video plot out of nothing.\\
        Begin!\\
        Video contexts in temporal order: \VarSty{\texttt{textualizing\_videos}}\\
        Question: \VarSty{\texttt{question}}
        \end{minipage}
        \end{tabular}
    \end{tcolorbox}
    \vspace{-2mm}
    \caption{\textbf{System prompt.} It allows the LLM to understand textualizing videos and respond according to relevant content in the document, avoiding answering questions that do not relate to the video.  }
    \label{tab:system_prompt}
    \vspace{-2mm}
\end{minipage}
\end{table*}

\paragraph{Analysis} Lite perception models enable {\modelname}-Text to convert videos into time-stamped text at 1 FPS, processing a 10-second video clip in about 2 seconds using an NVIDIA-A10 GPU. It communicates with users through an LLM. However, using text as the communication medium restricts the representation capabilities of the perception models, as it limits their decoders. To provide richer visual information from videos to the LLM, we must employ more advanced and more perception models, which may conflict with {\modelname}-Text's efficiency. Additionally, {\modelname}-Text has limited potential to benefit from popular visual instruction tuning \cite{llava}.

\subsection{\modelname-Embed\logoblue\ \ \ \  : {\modelname} by Encoding Videos as Embeddings}
\label{sec:videochat_embed}

{\modelname}-Embed is an end-to-end model designed to handle video-based dialogue. It employs an architecture that combines both video and language foundation models with an addition learnable Video-Language Token Interface (VLTF). To achieve better cross-modality optimization, the model incorporates language-friendly video foundation models, inspired by \cite{florence,videoclip,wang2022internvideo,umt}. Considering the video redundancy \cite{videomae}, we introduce the VLTF, using cross-attention to compress the video tokens. It is tuned with video-text data for video-to-language representation alignment. Finally, the video tokens, user queries, and dialogue context are input into the LLM for communication.

\begin{figure*}[h!]
    \centering
    \begin{minipage}{0.62\columnwidth}\vspace{0mm}    
        \includegraphics[width=1\textwidth]{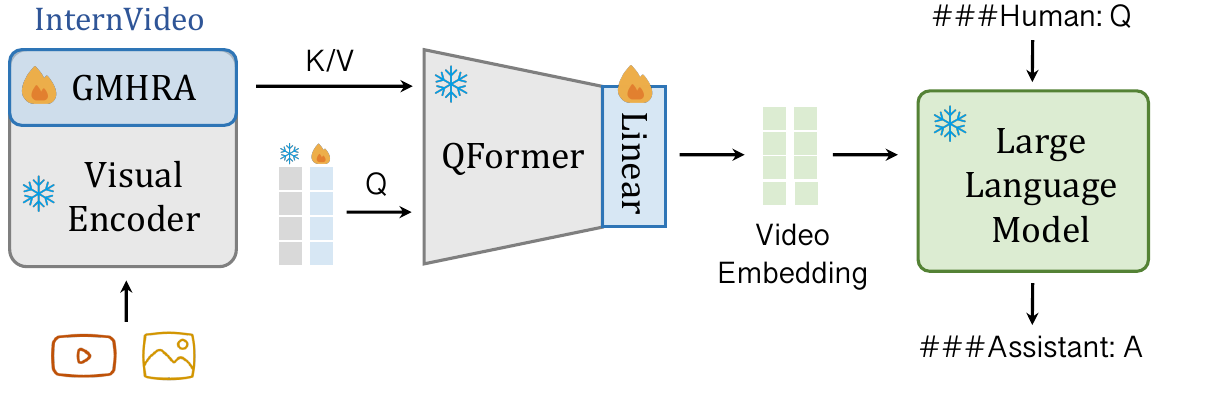}
        \vspace{-0.4cm}
        \subcaption{Architecture}
        \label{fig:video_embed}
    \end{minipage}
    \begin{minipage}{0.36\columnwidth}\vspace{0mm}    
        \includegraphics[width=1\textwidth]{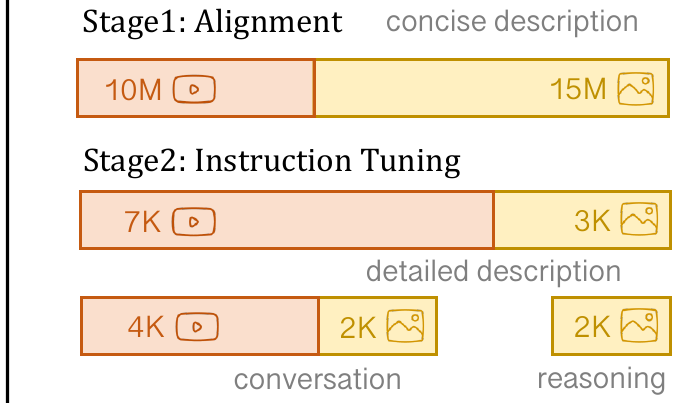}
        \vspace{-0.4cm}
        \subcaption{Data}
        \label{fig:instruct_data}
    \end{minipage}
    \vspace{-0.1cm}
    \caption{
    \textbf{Architecture and training paradigm of VideoChat-Embed.} It is built on BLIP-2 \cite{li2023blip} and StableVicuna \cite{stablelm}.
    The training contains two-stage alignment and instruction tuning.
    }
\end{figure*}

\subsubsection{Architecture}
In this paper, 
we instantiate the {\modelname}-Embed based on BLIP-2 \cite{li2023blip} and StableVicuna \cite{stablelm}(Figure \ref{fig:video_embed}). 
Concretely, we incorporate the pretrained ViT-G \cite{sun2023eva} with Global Multi-Head Relation Aggregator (GMHRA), a temporal modeling module used in InternVideo \cite{wang2022internvideo} and UniFormerV2 \cite{uniformerv2}. For the token interface, we employ the pretrained QFormer with extra linear projection, supplemented by additional query tokens to account for video context modeling. This allows us to obtain compact LLM-compatible video embeddings for dialogues.

When training, we freeze most of the parameters except the newly incorporated GMHRA, queries and linear projection. Inspired by \cite{umt}, we introduce image data for joint training (Figure \ref{fig:instruct_data}). In Stage1, we align the video encoder with LLM via large-scale video-text fine-tuning. In Stage2,
we tune the system with two types of video instruction data: in-depth video descriptions and video question-answer pairs. The following section will describe the process of generating instruction data and present the details of two-stage training paradigm.

\subsubsection{Instruction Data}
\label{sec:instruct_data}
We build a video-centric multimodal instruction data based on WebVid-10M \cite{webvid}. 
The corresponding detailed descriptions and question-answer generations are produced by ChatGPT based on video text (aided by {\modelname}-Text) with several prompts concerning spatiotemporal features. 
Compared with detailed video descriptions, video conversations are introduced to further improve the diversity, temporal and casual features in the video instruction data.

\paragraph{Detailed Video Descriptions} We condense the provided video description into a video narrative employing GPT-4, as shown in Table~\ref{tab:detailed_desc}. 
This highlights the temporal aspects of the video by illustrating its progression over time. 
The associated prompts can be found in Table \ref{tab:prompt_video_des} and \ref{tab:prompt_post}. 
The first  converts the various predicted textual labels into a cohesive, evolving story, while the second one refines the narrative to improve clarity and coherence, as well as minimize hallucination. 
We generated a total of 7K descriptions from randomly chosen videos.

\begin{table*}[t]\centering
    \begin{minipage}{0.99\columnwidth}\vspace{0mm}    
    \centering
    \begin{tcolorbox} 
        \centering
        \hspace{-6mm}
        \begin{tabular}{p{0.99\columnwidth}}
        \hspace{1mm}
        \begin{minipage}{0.99\columnwidth}
        Give you a video of \VarSty{\texttt{origin\_caption}}. The content of the video in temporal order is: \VarSty{\texttt{textualizing\_videos}}. Please use the sequence adverbs "first", "next", "then" and "finally" to describe this video in detail, but don't mention the specific time. Give as many details as possible. Say everything you see. The description should be more than 150 words and less than 200 words. 
    \end{minipage}
        \end{tabular}
    \end{tcolorbox}
    \vspace{-2mm}
    \caption{\textbf{Prompts for detailed video descriptions.} The \VarSty{\texttt{origin\_caption}} is generated from \modelname-Text.}
    \label{tab:prompt_video_des}
    \end{minipage}
    \vspace{2mm}
    
    \begin{minipage}{0.99\columnwidth}\vspace{0mm}    
    \centering
    \begin{tcolorbox} 
        \centering
        \hspace{-6mm}
        \begin{tabular}{p{0.99\columnwidth}}
        \hspace{1mm}
        \begin{minipage}{0.99\columnwidth}
        Fix the error in the given paragraph. Remove any repeating sentences, meaningless characters, not English sentences, and so on. Remove unnecessary repetition. Rewrite any incomplete sentences. Return directly the results without explanation. Return directly the input paragraph if it is already correct without explanation.
    \end{minipage}
        \end{tabular}
    \end{tcolorbox}
    \vspace{-2mm}
    \caption{\textbf{Prompt for post-processing.} Copied from miniGPT-4 \cite{minigpt}.}
    \label{tab:prompt_post}
    \end{minipage}
    \vspace{2mm}
    
    \begin{minipage}{0.99\columnwidth}\vspace{0mm}    
    \centering
    \begin{tcolorbox} 
        \centering
        % \small
        \hspace{-6mm}
        \begin{tabular}{p{0.99\columnwidth}}
        \hspace{1mm}
        \begin{minipage}{0.99\columnwidth}
        \includegraphics[width=1\textwidth]{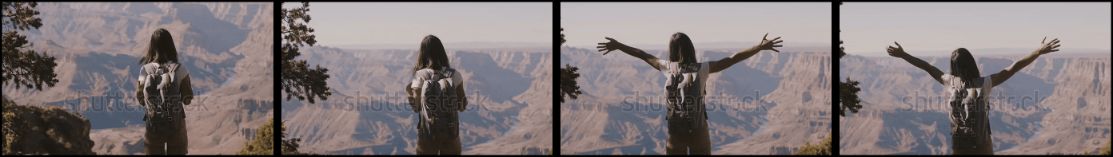}\\
        \texttt{\#\#\#Human:} \texttt{<Video>\VarSty{video\_embed}</Video>} The video contains \VarSty{T} frames sampled at \VarSty{$t_{0}, t_{1}, ... ,t_{T}$} seconds. \\
        \texttt{\#\#\#Human:} Portray the video with a rich, descriptive narrative, capturing its atmosphere and events. \\
        \texttt{\#\#\#Assistant:} \textcolor{red}{First}, we see a young woman with long brown hair wearing a backpack standing on the edge of a cliff, looking at the beautiful mountain scenery. She is in awe and stands there for a few seconds. \textcolor{red}{Next}, we see her turning around slowly, with her arms open wide. \red{Then}, we see her taking a few steps back, still facing the canyon, with her arms still open wide. She seems to be soaking in the grandeur of the view. \textcolor{red}{Finally}, she stands confidently at the edge of the cliff, her backpack still on her back, as if she has conquered the mountain. Throughout the video, we see the vast expanse of the canyon with the mountains in the background. The sky is overcast in some parts and clear blue in others, creating a beautiful contrast. \textcolor{red}{Overall}, the young woman seems to be thrilled with the view and enjoys every bit of it with open arms. 
        \end{minipage}
        \end{tabular}
    \end{tcolorbox}
    \vspace{-2mm}
    \caption{\textbf{Example of using the detailed video description.}}
    \label{tab:detailed_desc}
    \end{minipage}
    \vspace{-2mm}
\end{table*}

\begin{table*}[h!]\centering
    \begin{minipage}{0.99\columnwidth}\vspace{0mm}    
    \centering
    \begin{tcolorbox} 
        \centering
        \hspace{-6mm}
        \begin{tabular}{p{0.99\columnwidth}}
        \hspace{1mm}
        \begin{minipage}{0.99\columnwidth}
    As an AI visual assistant, you are observing a single video. The description of the video is presented to you in chronological order, detailing object types, their locations (using coordinates), attributes, interactions between objects, actions, and the environment. Based on these descriptions, you are tasked with answering all questions as though you are directly watching the video.

Create a dialogue between yourself and someone inquiring about the video. Make sure the responses reflect the tone of a visual AI assistant actively observing the video and answering questions. Include diverse queries and corresponding answers.

Incorporate \textcolor{cyan}{questions that address the visual content of the video, such as object types, attributes, object counting, actions, locations, relative positions between objects, and changes in object actions or locations over time, as well as object interactions.} Only include questions with definitive answers:

$\bullet$ Questions whose contents can be confidently observed and answered based on the video.
$\bullet$ Questions whose absence from the video can be confidently determined.

Next, encompass \textcolor{blue}{questions related to temporal perception and reasoning, such as inquiring about what a person did before or after an event, or asking for specific timestamps of certain events or actions}

Also include \textcolor{purple}{complex questions relevant to the video's content, like those asking about the background knowledge of objects or actions in the video, discussing events occurring in the video, delving into counterfactual topics (e.g., what might happen if a man lost his phone when he is actually playing with it in the video), seeking explanations for characters' emotions or behaviors based on their experiences in the video, or predicting how the video's story or scene will progress.}

Since you receive video descriptions while viewing the video, \textbf{prioritize asking more questions about visual changes over time and the reasons or causes behind these changes rather than questions that can be inferred from a single frame.}

Remember not to inquire about uncertain details. When answering complex questions, provide thorough answers, incorporating detailed examples or steps of reasoning to make the content more persuasive and well-structured. Use multiple paragraphs if necessary. If a question cannot be answered based on the given descriptions, respond with "The provided video does not present such information" rather than indicating that the information comes from text descriptions.
 \end{minipage}
        \end{tabular}
    \end{tcolorbox}
    \vspace{-2mm}
    \caption{\textbf{Prompt for video conversations.} The \textcolor{cyan}{descriptive}, \textcolor{blue}{temporal}, and \textcolor{purple}{causal} question specifications are marked by \textcolor{cyan}{cyan}, \textcolor{blue}{blue}, and \textcolor{purple}{purple}, respectively. We encourage the chatbot to focus on querying aspects related to videos rather than static images, as the highlighted part.}
    \label{tab:prompt_video_qa}
\end{minipage}
    \vspace{-2mm}
\end{table*}

\begin{table*}[h!]\centering
\begin{minipage}{0.99\columnwidth}\vspace{0mm}    
    \centering
    \begin{tcolorbox} 
        \centering
        % \small
        \hspace{-6mm}
        \begin{tabular}{p{0.99\columnwidth}}
        \hspace{1mm}
        \begin{minipage}{0.99\columnwidth}
        \includegraphics[width=1\textwidth]{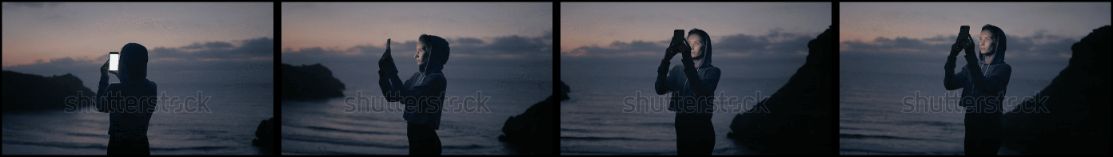}\\
        \texttt{\#\#\#Human:} \texttt{<Video>\VarSty{video\_embed}</Video>} The video contains \VarSty{T} frames sampled at \VarSty{$t_{0}, t_{1}, ... ,t_{T}$} seconds. \\
        \texttt{\#\#\#Human:} What is happening in the video? \\
        \texttt{\#\#\#Assistant:} In the video, we see a silhouette of a young woman holding a smartphone and taking pictures of the sunset over the sea on the beach at night. Then, there are several other silhouettes of people taking pictures and looking at their phones. \\
        \texttt{\#\#\#Human:} Can you describe the \red{interactions} between the objects in the video? \\
        \texttt{\#\#\#Assistant:} The silhouettes of people in the video are holding smartphones, and she is taking pictures of the sunset over the sea on the beach at night. \\
        \texttt{\#\#\#Human:} What is the \red{location} of the scene? \\
        \texttt{\#\#\#Assistant:} The scene is located on a beach at night near the sea. \\
        \texttt{\#\#\#Human:} Can you tell me about any changes that happen in the video \red{over time}? \\
        \texttt{\#\#\#Assistant:} The color of the sky changes from orange to grey as the video progresses. Additionally, the actions of the silhouettes of people change as they take pictures and look at their phones on the beach.
        \end{minipage}
        \end{tabular}
    \end{tcolorbox}
    \vspace{-2mm}
    \caption{\textbf{Example of using the video conversation.}}
    \label{tab:multi_qa}
    \end{minipage}
    \vspace{-2mm}
\end{table*}

\paragraph{Video Conversations}
With the video description, we generate multi-round dialogues with three types of prompts concerning descriptive, temporal, and causal content for videos with ChatGPT. The descriptive part mostly inherits key points from LLaVA~\cite{llava}. For the temporal and causal parts, we propose prompts (Table \ref{tab:prompt_video_qa}) focus on temporal perception/reasoning and explanation/uncovering intentions/causes, respectively. We produced multi-round dialogues from 4K randomly chosen videos.
One example of the video conversation can be found in Table \ref{tab:multi_qa}

\subsubsection{Two-stage Training}
Motived by MiniGPT-4 \cite{minigpt} and LLaVA \cite{llava},
we have designed a two-stage joint training paradigm. This approach allows us to benefit from readily-available image instruction data, creating a system capable of handling both images and videos with shared spatial perception and reasoning capacity.

\paragraph{Stage1: Alignment.}
To strike a balance between training convergency and efficiency
we introduce 25M vision-text pairs for one epoch of fine-tuning,
The data consists 10M video-text pairs from WebVid-10M,
and 15M image-text pairs from COCO Caption~\cite{cococaptions}, Visual Genome~\cite{vg}, SBU Captions~\cite{sbu}, CC3M~\cite{cc3m} and CC12M~\cite{cc12m}.
The input prompts for LLMs are as followed:

\begin{itemize}[leftmargin=6mm]
    \setlength{\itemsep}{2pt}
    \item ``\texttt{\#\#\#Human:} \texttt{<Video>\VarSty{video\_embed}</Video> \VarSty{video\_instruction} \#\#\#Assistant:}''
    \item ``\texttt{\#\#\#Human:} \texttt{<Image>\VarSty{image\_embed}</Image> \VarSty{image\_instruction} \#\#\#Assistant:}''
\end{itemize}

The \VarSty{video\_embed} and \VarSty{image\_embed} are the output from the token interface. Meanwhile the \VarSty{video\_instruction} and \VarSty{image\_instruction} provide concise video and image descriptions randomly sampled from predefined instructions in Table \ref{tab:concise_describe_instructions}. Language models receive corresponding visual descriptions as answers.

\paragraph{Stage2: Instruction tuning.}
As discussed in Section \ref{sec:instruct_data}, our self-built video instruction data consists of 7K detailed video descriptions and 4K video conversations. To improve spatial perception and reasoning capabilities, we also gather 3K detailed image descriptions from MiniGPT-4 \cite{minigpt}, 2K image conversations, and 2K image reasoning tasks from LLaVA \cite{llava}. With this 18K data collection, we tune the system for 3 epochs. Note we include temporal reasoning sampling information for video data: ``The video contains \VarSty{T} frames sampled at \VarSty{$t_{0}, t_{1}, ... ,t_{T}$} seconds.''

\section{Experiments}

\subsection{Qualitative Analysis}
\begin{figure}[htbp]
\centering
\begin{minipage}[b]{0.48\linewidth}
\centering
\includegraphics[width=\linewidth]{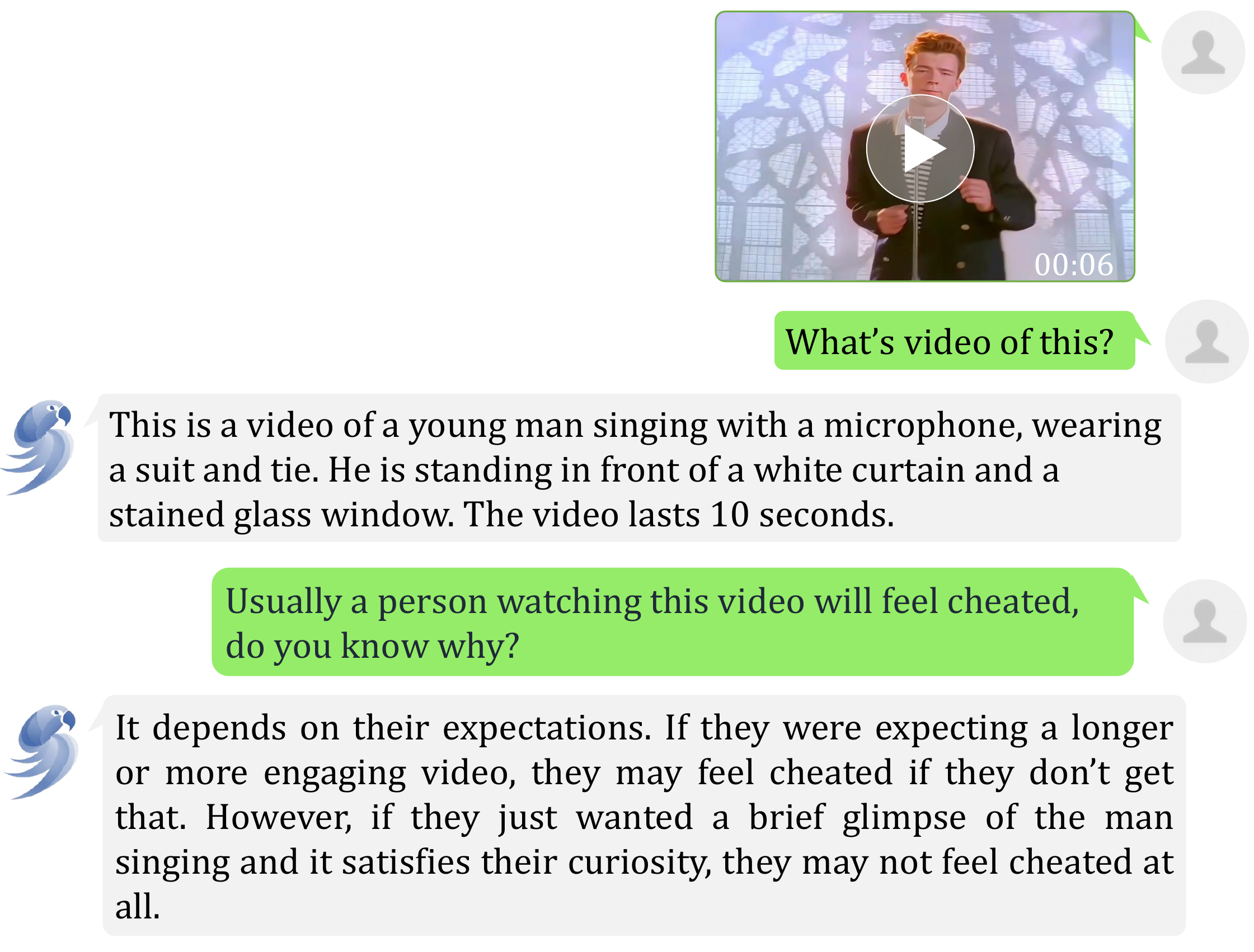}
\caption{Video meme case. ~\href{https://pjlab-gvm-data.oss-cn-shanghai.aliyuncs.com/papers/media/jesse_dance.mp4}{Click here to watch the video.}}
\label{fig:roll}
\end{minipage}
\hspace{3mm}
\begin{minipage}[b]{0.48\linewidth}
\centering
\includegraphics[width=\linewidth]{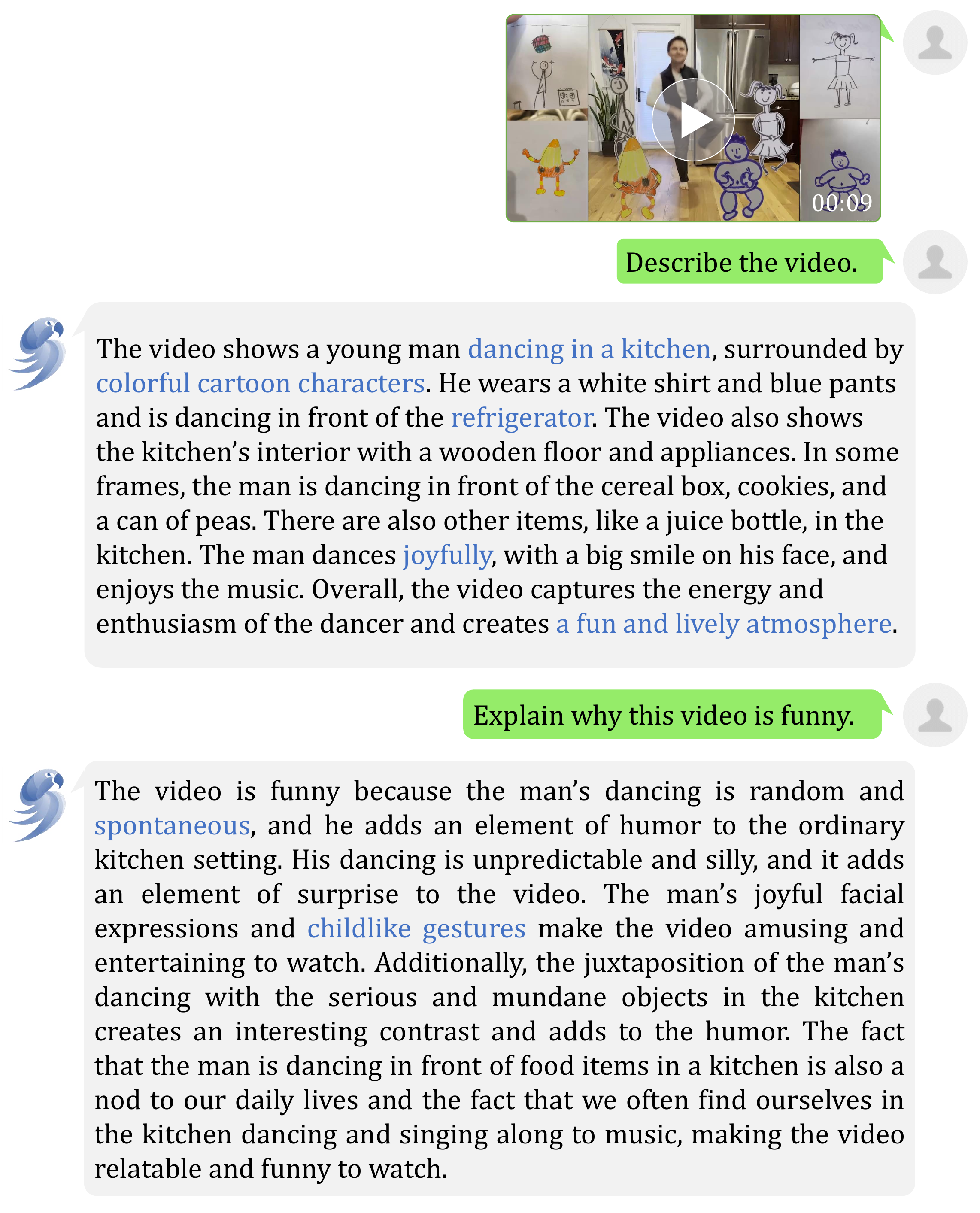}
\caption{\modelname~can understand why the video is funny. ~\href{https://pjlab-gvm-data.oss-cn-shanghai.aliyuncs.com/papers/media/jesse_dance.mp4}{Click here to watch the video.}}
\label{fig:funny_video}
\end{minipage}
\end{figure}

\begin{figure}[htbp]
\centering
\begin{minipage}[b]{0.48\linewidth}
\centering
\includegraphics[width=\linewidth]{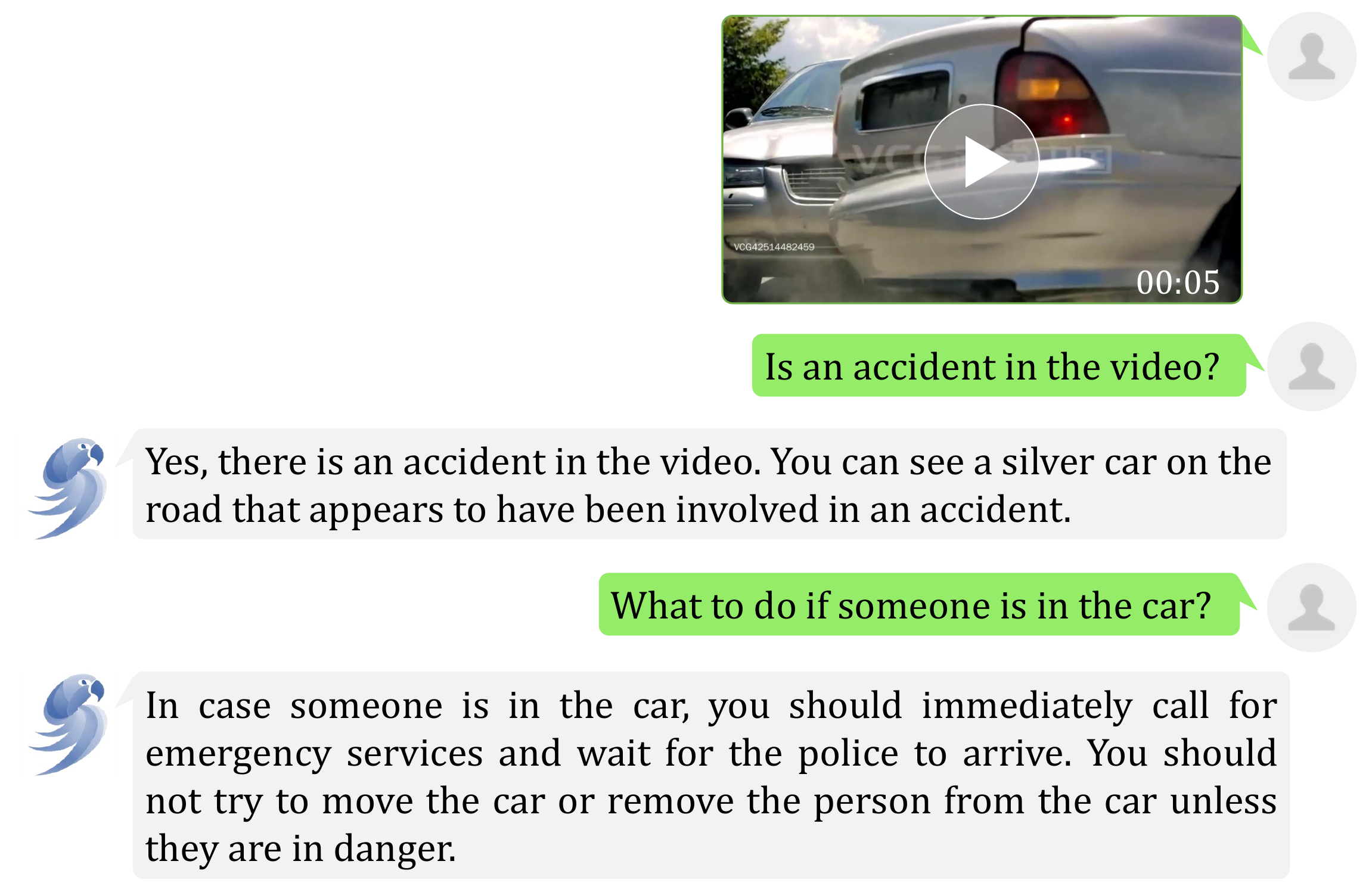}
\caption{\modelname~finds the accident in the video and gives the advice to the person in the car.~\href{https://pjlab-gvm-data.oss-cn-shanghai.aliyuncs.com/papers/media/car_accident.mp4}{Click here to watch the video.}}
\label{fig:accident}
\end{minipage}
\hspace{3mm}
\begin{minipage}[b]{0.48\linewidth}
\centering
\includegraphics[width=\linewidth]{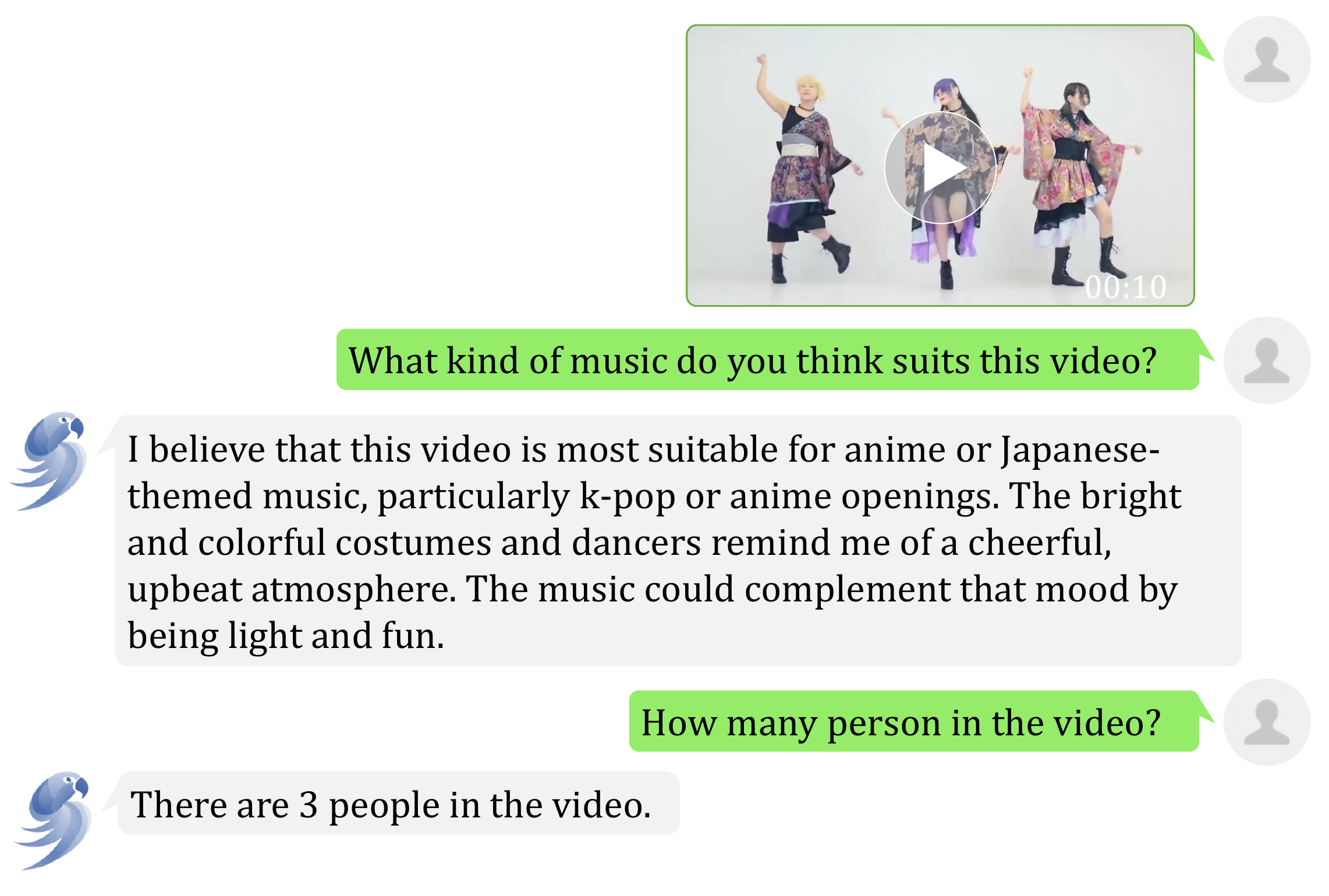}
\caption{\modelname~observed the number of people and their clothing styles in the space, and perceived temporal actions to recommend light and funny music.~\href{https://pjlab-gvm-data.oss-cn-shanghai.aliyuncs.com/papers/media/jp_dance.mp4}{Click here to watch the video.}}
\label{fig:dance}
\end{minipage}
\end{figure}

\begin{figure}[htbp]
\centering
\begin{minipage}[b]{0.48\linewidth}
\centering
\includegraphics[width=\linewidth]{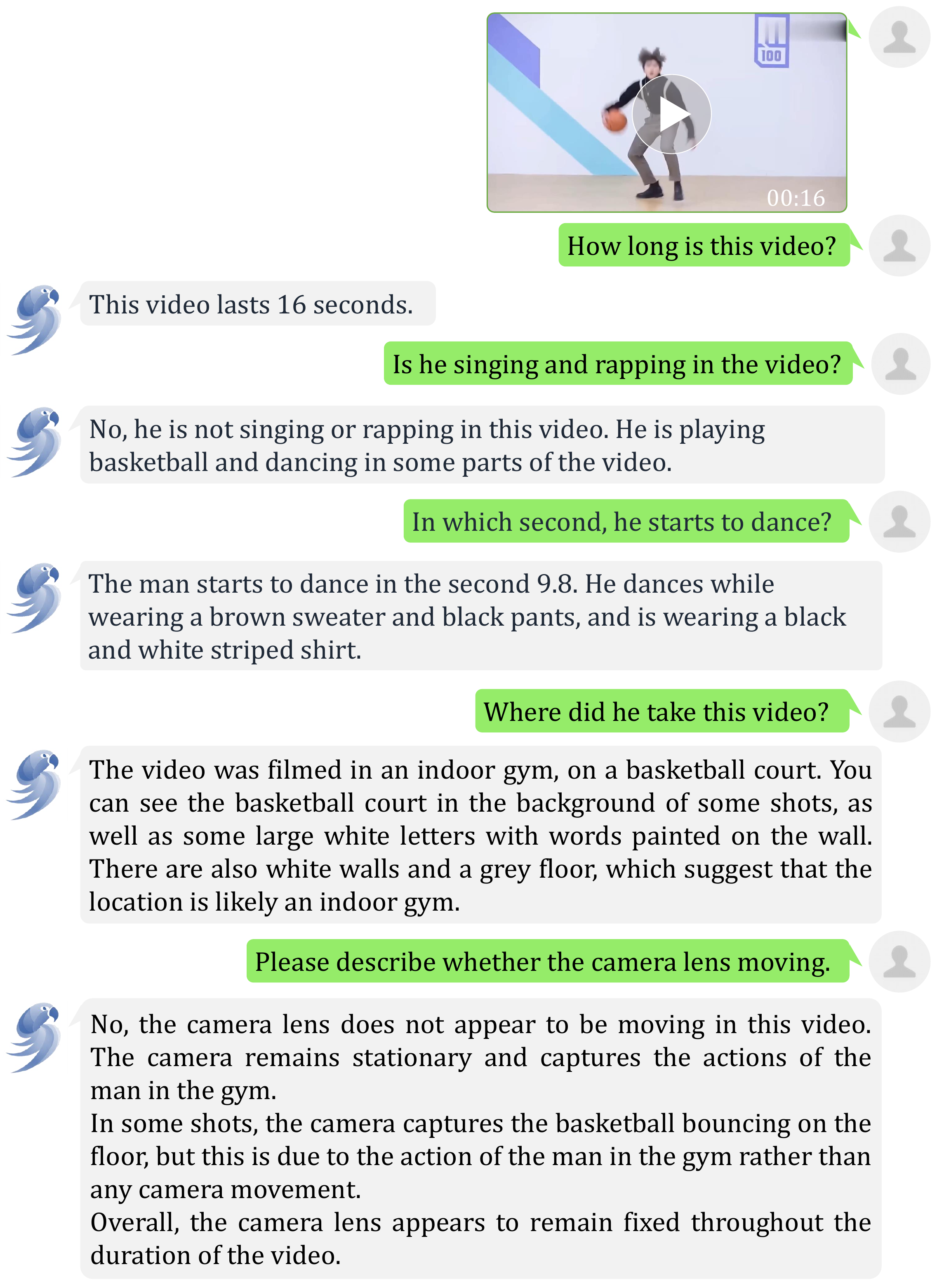}
\caption{Multi-turn conversation with VideoChat-Embed.~\href{https://pjlab-gvm-data.oss-cn-shanghai.aliyuncs.com/papers/media/idol_dancing.mp4}{Click here to watch the video.}}
\label{fig:idol}
\end{minipage}
\hspace{3mm}
\begin{minipage}[b]{0.48\linewidth}
\centering
\includegraphics[width=\linewidth]{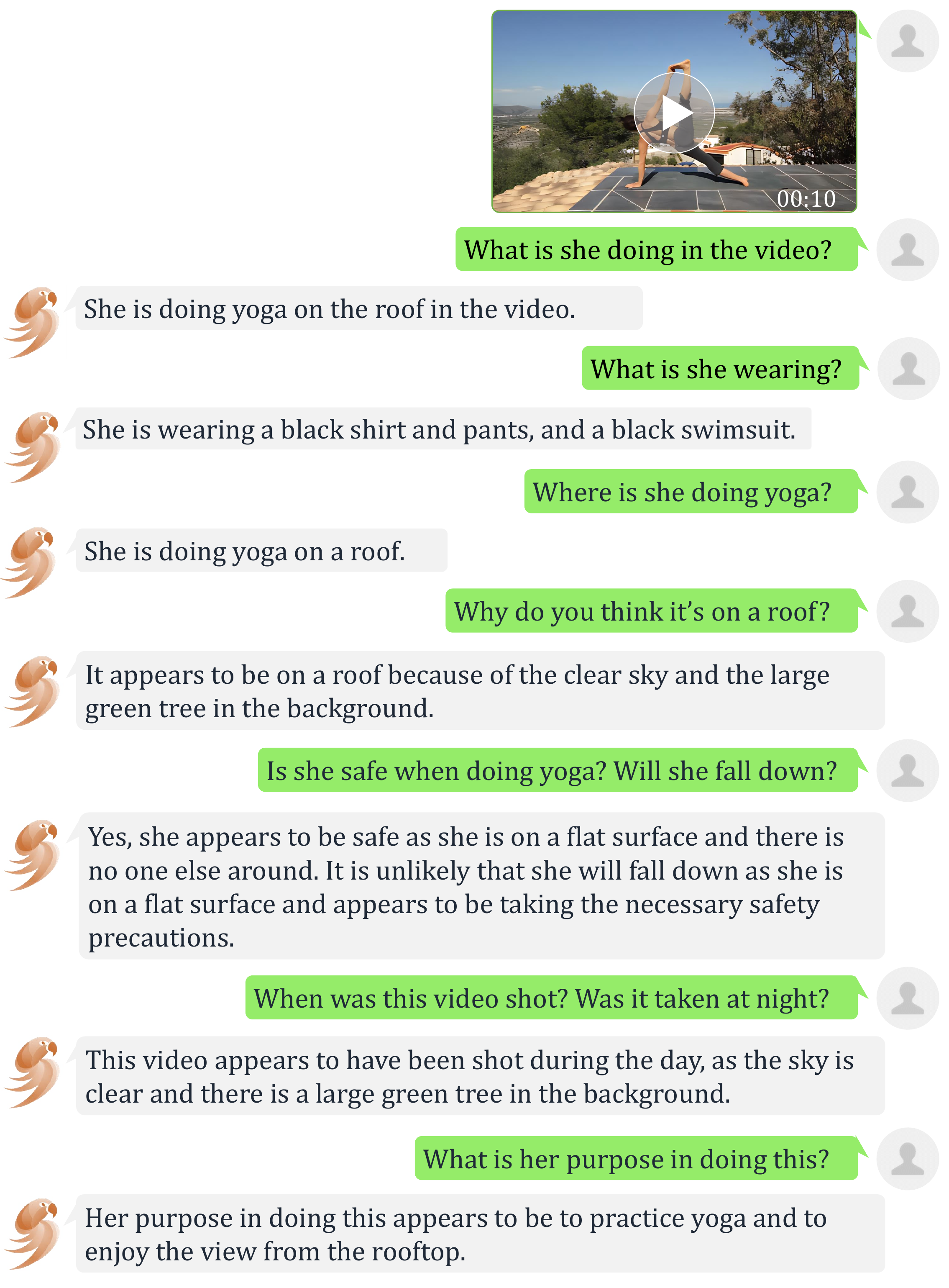}
\caption{Multi-turn conversation with VideoChat-Text.~\href{https://pjlab-gvm-data.oss-cn-shanghai.aliyuncs.com/papers/media/yoga.mp4}{Click here to watch the video.}}\
\label{fig:yoga}
\end{minipage}
\end{figure}

\begin{figure}[htbp]
\centering
\begin{minipage}[b]{0.32\linewidth}
\centering
\includegraphics[width=\linewidth]{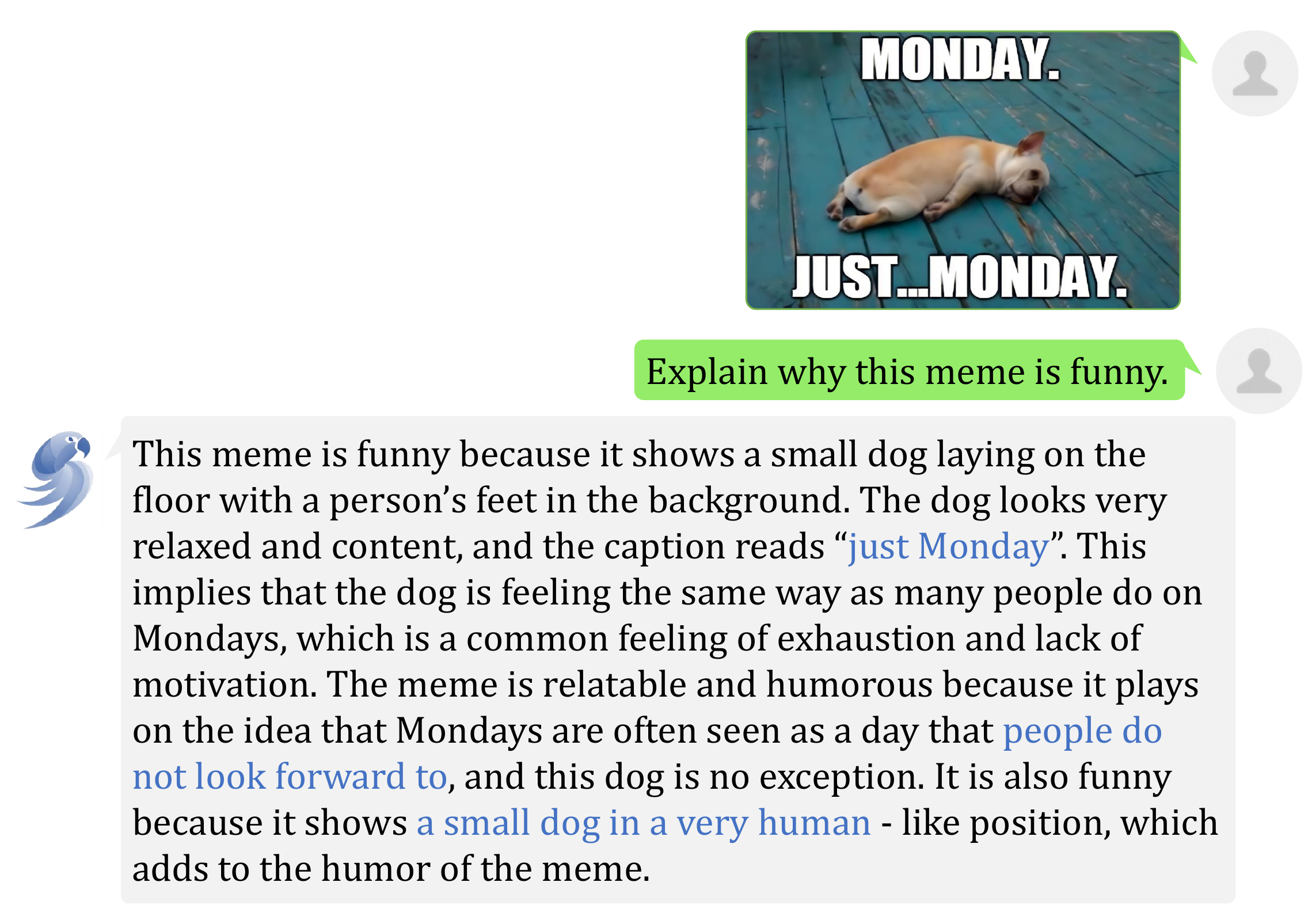}
\end{minipage}
\begin{minipage}[b]{0.32\linewidth}
\centering
\includegraphics[width=\linewidth]{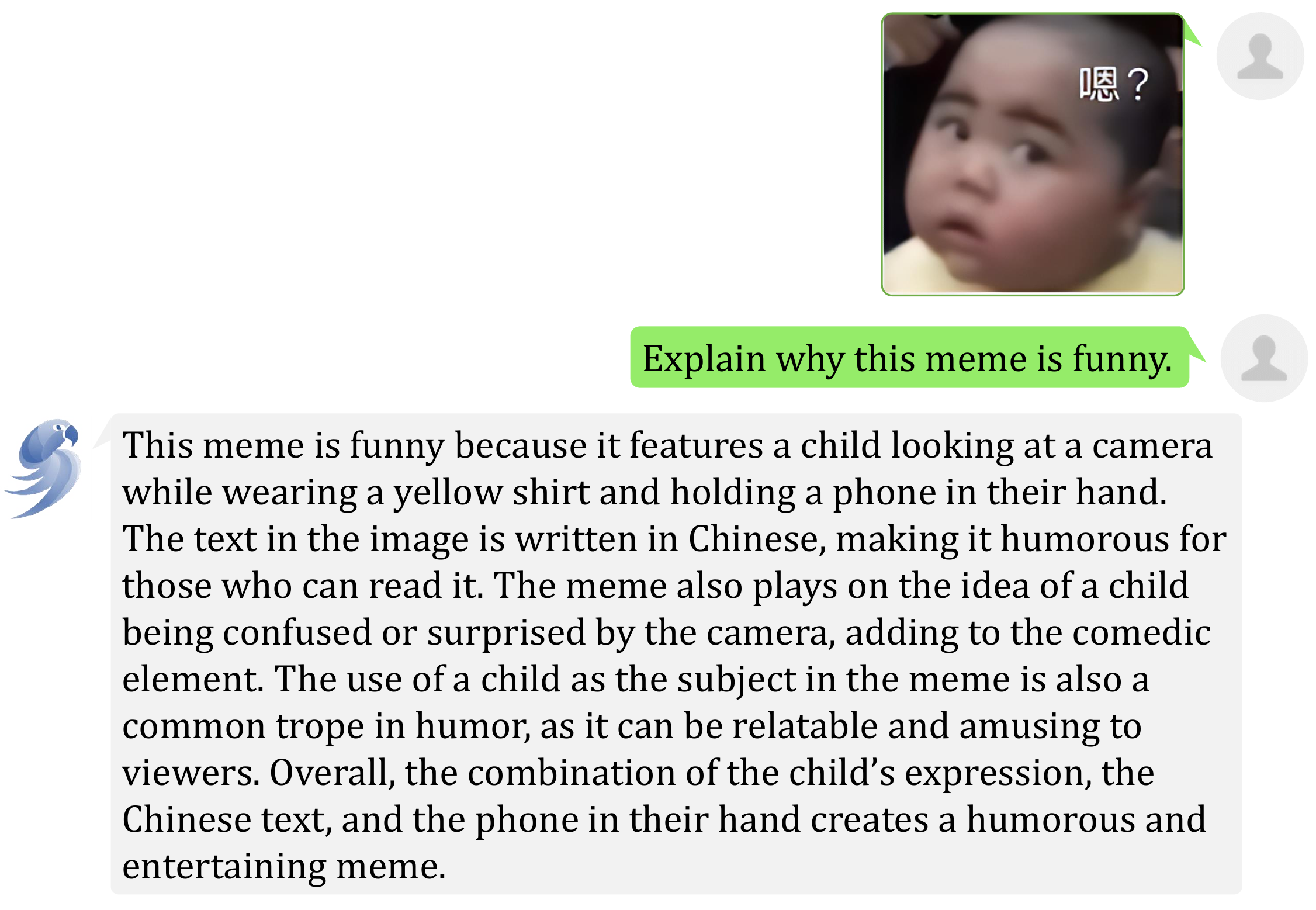}
\end{minipage}
\begin{minipage}[b]{0.32\linewidth}
\centering
\includegraphics[width=\linewidth]{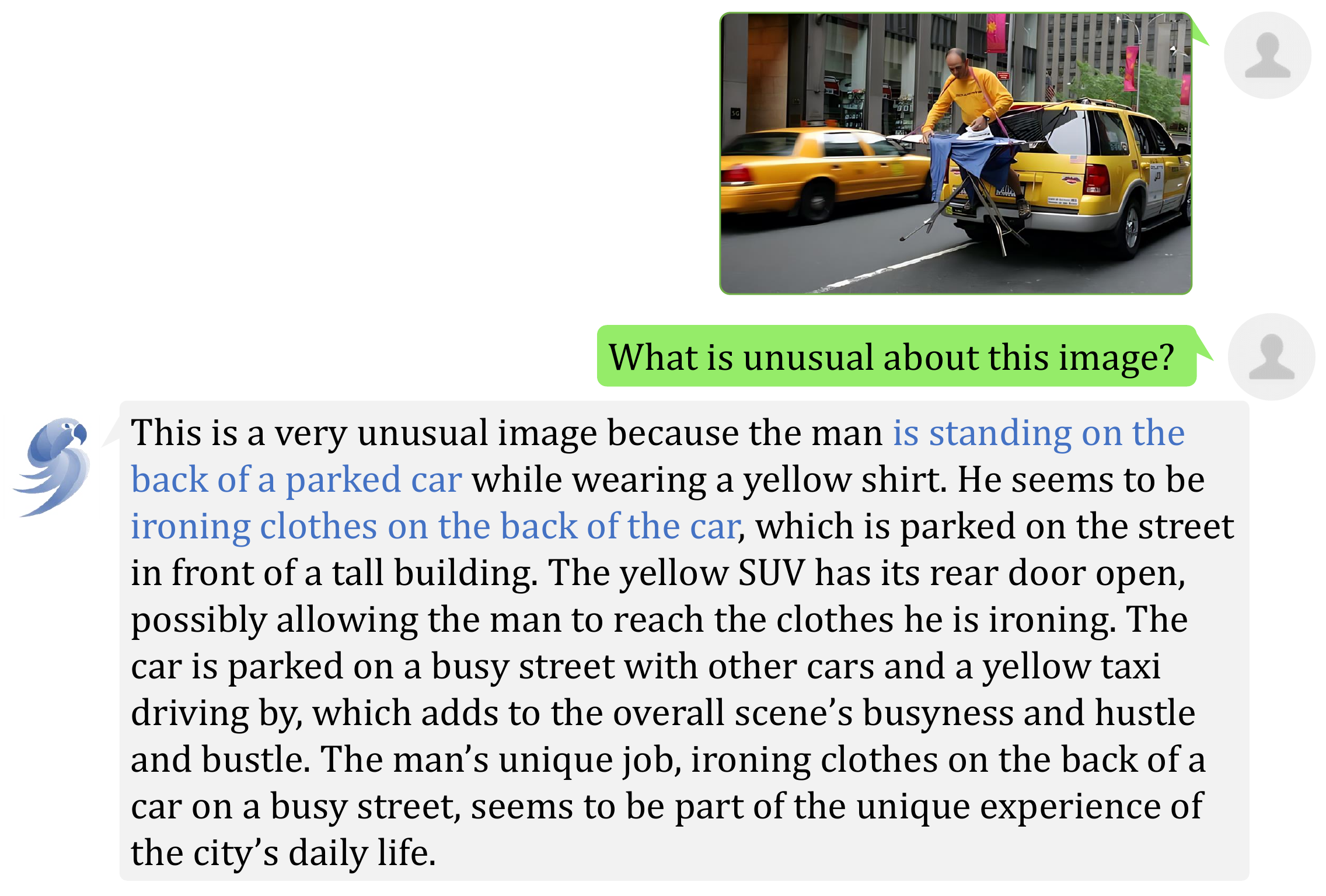}
\end{minipage}
\caption{Cases of Image Comprehension. Please zoom in for details.}
\label{fig:three_pics2}
\end{figure}

We give some case studies at this stage. Besides of our {\modelname}-Text\logotext\ \ \ \ and {\modelname}-Embed\logoblue\ \ \ \ , we make qualitative comparisons with LLaVa\logollava\ \  ~\cite{llava}, miniGPT-4\logominigpt\ \ \  ~\cite{minigpt}, and mPLUG-owl\logoowl\ \  ~\cite{mplugowl}.
\paragraph{Spatial Perception and Analysis} 
In Figure~\ref{fig:dance}, our approach ({\modelname}-Embed) accurately deduces the corresponding music by recognizing Japanese-style clothing and determining the number of individuals present. This confirms the system's ability to identify objects along with their properties, while also providing pertinent recommendations based on visual elements. Also, we give some image-centric dialogue examples in Figure \ref{fig:three_pics2}.
\paragraph{Temporal Perception and Reasoning} Figure~\ref{fig:funny_video}, \ref{fig:dance}, and~\ref{fig:yoga} demonstrate that {\modelname}-Embed is capable of performing accurate temporal perception and reasoning. In Figure~\ref{fig:funny_video}, our system identifies actions over time in a zero-shot fashion, recognizing that the subject played basketball and engaged in dance movements within a specific timeframe. Additionally, it captures camera motion, showcasing its understanding of filming perspectives. In Figure~\ref{fig:yoga}, {\modelname}-Text accurately identifies yoga in the video and provides rational explanations for this activity (practice and enjoyment). Intriguingly, when questioned about the likelihood of the yoga practitioner falling, it asserts that proper safety precautions were taken by the individual.
\paragraph{Casual Inference} It is evident that {\modelname}-Embed can infer causal relationships using spatiotemporal clues, as demonstrated in Figure \ref{fig:funny_video}, \ref{fig:accident}, \ref{fig:dance}, and~\ref{fig:roll}. In Figure \ref{fig:funny_video}, the model provides an impartial description of the video, primarily highlighting objects, actions, and emotions without commenting on the boy's dance style. To explain why the video is amusing, {\modelname}-Embed cites the erratic and spontaneous nature of the boy's movements while also conveying an emotional assessment that his dancing appears foolish, accounting for the unexpected humor within the clip. Empirically, we confirm that these visually associated abstract concepts are derived from the video foundation model instead of being hallucinations produced by the utilized LLM. In Figure \ref{fig:accident}, {\modelname}-Embed accurately assesses that a car accident occurred due to the collision of vehicles and the damage sustained to the front car's license plate visible to the camera. In Figure \ref{fig:dance}, {\modelname} suggests pairing the video with light, enjoyable music, as it can sense the girls' dancing rhythm and progression over time. This demonstrates that our system can effectively capture and summarize abstract concepts related to movement patterns. 

\subsection{Comparisons} 
\begin{figure}
    \centering
    \includegraphics[width=\linewidth]{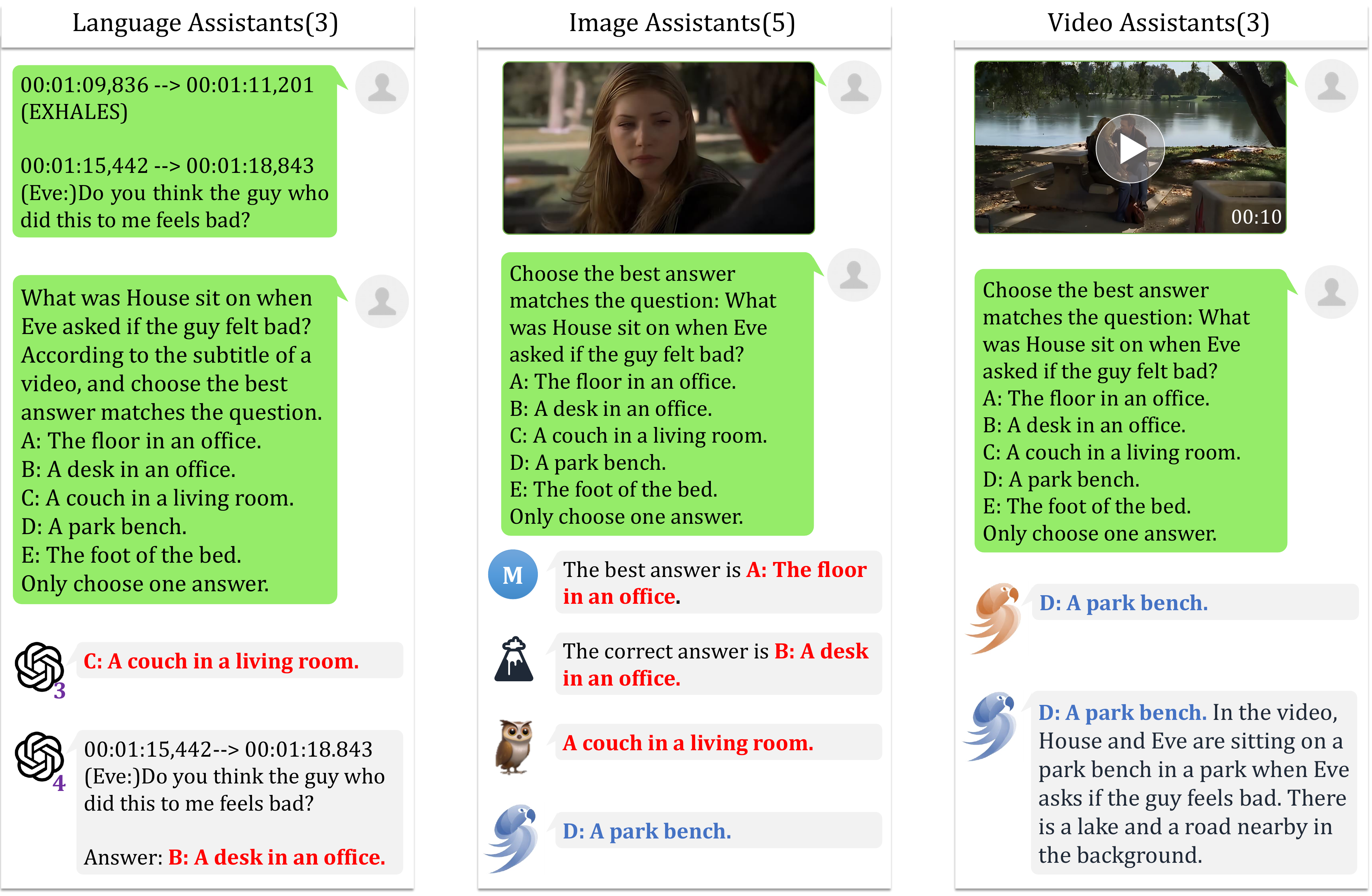}
    \caption{In the TVQA dataset sample, subtitles are fed to language assistants~\cite{openai2022chatgpt,openai2023gpt4}, images to the image assistants~\cite{zhu2023minigpt,llava,mplugowl}, and videos to the {\modelname}. While responding to user questions, only {\modelname} utilizes information from videos.}
    \label{fig:my_label}
\end{figure}

As depicted in Figure~\ref{fig:my_label}, we present a comparison of our approach to recent image-based multimodal dialogue systems in image-related tasks, using a query example from the TVQA dataset. We assess this case through the online demos provided by each respective system. It is evident that our {\modelname}-Embed correctly identifies the scene, while other systems inaccurately perceive the conversation setting as indoors. This result highlights the superior spatial perception abilities of {\modelname}-Embed in relation to its counterparts. Furthermore, this proficiency remains consistent when dealing with a video from the same dataset, as demonstrated in the right-hand example.

\section{Conclusion}
We have embarked on a pioneering investigation into general video understanding by developing VideoChat, a multimodal dialogue system specifically designed for videos. 
Two versions of VideoChat are implemented: a text-based version demonstrates the effectiveness of employing large language models as universal decoders for video tasks, while an end-to-end version makes a preliminary attempt to tackle video understanding using instructed video-to-text formulation. Our end-to-end solution effectively merges video foundation models with large language models through a trainable neural interface. To enhance the system's performance, we introduced a video-centric instructional dataset, which highlights spatiotemporal reasoning and causality by offering a learning resource for video-based multimodal dialogue systems. Initial qualitative evaluations showcase our system's promising capabilities across various video applications and drive its ongoing advancement.

\paragraph{Limitations}
1) Both {\modelname}-Text and {\modelname}-Embed  struggle with managing long-term videos ($\ge$ 1min). On one side, effectively and efficiently modeling the context of long videos remains a complex and persistent research issue. Conversely, balancing response time, GPU memory usage, and user expectations for system performance becomes challenging when striving to provide user-friendly interactions while processing longer videos.
2) Our system's capacities for temporal and causal reasoning remain rudimentary. These limitations stem from the current scale of our instruction data and its construction approaches, as well as the overall system scale and the models employed.
3) Addressing performance disparities in time-sensitive and performance-critical applications, such as egocentric task instruction prediction and intelligent monitoring, is an ongoing challenge.

\paragraph{Future Works}
Our future works lie in 1) scaling video foundation models both in capacity and data for better spatiotemporal modeling, 2) video-centric multimodal training data and reasoning benchmark for evaluations at scale, and 3) long-term video processing techniques.

% 1. Introduction
% First good try, enhancing video-language understanding
% 2. Related work: 
% (1) Video-language understanding
% (2) Large Language Models
% (3) LLMs for Multimodal Understanding
%     Our work design two styles of VideoChat: Explicit and Implicit
% 3. Emplicit VideoChat
% (1) Framework: light color for parrot logo
% (2) Tools: InternVideo, Tag2Text, Grid, T5, Whisper, Langchain
% (3) System (prompt design)
% 4. Implicit VideoChat
% (1) Framework: dark color for parrot logo
% (2) Instruction data: ours, minigpt4, llava
% (3) Two-stage training: prompt, hyperparameters
% 5. Experiments
% (1) Easy Cases
% (2) Hard cases:
%    (i) explicit vs. implicit
%    (ii) comparison with text-only GPT-4 (TVQA)
%    (iii) counting, tal, reasoning
%    (iv) long-term
% (3) Discussion: Strength, weakness
% 6. Conclusion
% (1) conclusion
% (2) simple limitation
% (3) simple future work

%\section{Acknowledgement}
{\small
\bibliographystyle{plain}
\bibliography{egbib}
}

\newpage
\appendix
\section{Appendix}

\paragraph{Instruction for a brief description.}
Following the brief image instruction in LLaVA,
we generate the video instruction with the aid of ChatGPT as shown in Table \ref{tab:concise_describe_instructions}.
In Stage1,
we randomly sample the instruction to generate brief descriptions of images and videos.

\paragraph{Instruction for a detailed description.}
Following the detailed image instruction in LLaVA,
we generate the video instruction with the help of ChatGPT as shown in Table \ref{tab:detailed_describe_instructions}
To build the instruction data used in Stage2,
we randomly sample the instruction and combine it with the detailed descriptions.

\begin{table*}[h!]\centering
\begin{minipage}{0.99\columnwidth}\vspace{0mm}    
    \centering
    \begin{tcolorbox} 
        \centering
        \small
        \hspace{-6mm}
        \begin{itemize}[leftmargin=4mm]
        \setlength{\itemsep}{2pt}
            \item "Describe the following image concisely."
            \item "Provide a brief description of the given image."
            \item "Offer a succinct explanation of the picture presented."
            \item "Summarize the visual content of the following image."
            \item "Give a short and clear explanation of the subsequent image."
            \item "Share a concise interpretation of the image provided."
            \item "Present a compact description of the photo's key features."
            \item "Relay a brief, clear account of the picture shown."
            \item "Render a clear and concise summary of the photo below."
            \item "Write a terse but informative summary of the following picture."
            \item "Create a compact narrative representing the image presented."
        \end{itemize}
        \rule[0.25\baselineskip]{\textwidth}{1pt}
        \begin{itemize}[leftmargin=4mm]
        \setlength{\itemsep}{2pt}
            \item "Describe the following video concisely."
            \item "Provide a brief description of the given video clip."
            \item "Offer a succinct explanation of the footage presented."
            \item "Summarize the visual content of the following video."
            \item "Give a short and clear explanation of the subsequent video clip."
            \item "Share a concise interpretation of the video provided."
            \item "Present a compact description of the clip's key features."
            \item "Relay a brief, clear account of the video shown."
            \item "Render a clear and concise summary of the video below."
            \item "Write a terse but informative summary of the following video clip."
            \item "Create a compact narrative representing the video presented."
        \end{itemize}
    \end{tcolorbox}
    \vspace{-2mm}
    \caption{\textbf{The list of instructions for brief image and video description.}
    The image list is inherited from LLaVA \cite{llava}.
    The video list is generated by ChatGPT with examples of the image list.
    }
    \label{tab:concise_describe_instructions}
\end{minipage}
\end{table*}

\begin{table*}[h!]\centering
\begin{minipage}{0.99\columnwidth}\vspace{0mm}    
    \centering
    \begin{tcolorbox} 
        \centering
        \small
        \hspace{-6mm}
        \begin{itemize}[leftmargin=4mm]
        \setlength{\itemsep}{2pt}
            \item "Describe the following image in detail."
            \item "Provide a detailed description of the given image."
            \item "Give an elaborate explanation of the image you see."
            \item "Share a comprehensive rundown of the presented image."
            \item "Offer a thorough analysis of the image."
            \item "Explain the various aspects of the image before you."
            \item "Clarify the contents of the displayed image with great detail."
            \item "Characterize the image using a well-detailed description."
            \item "Break down the elements of the image in a detailed manner."
            \item "Walk through the important details of the image."
            \item "Portray the image with a rich, descriptive narrative."
            \item "Narrate the contents of the image with precision."
            \item "Analyze the image in a comprehensive and detailed manner."
            \item "Illustrate the image through a descriptive explanation."
            \item "Examine the image closely and share its details."
            \item "Write an exhaustive depiction of the given image."
        \end{itemize}
        \rule[0.25\baselineskip]{\textwidth}{1pt}
        \begin{itemize}[leftmargin=4mm]
        \setlength{\itemsep}{2pt}
            \item "Describe the following video in detail, including the actions and scenes."
            \item "Provide a detailed description of the given video, capturing its key moments."
            \item "Give an elaborate explanation of the video you see, including the events and characters."
            \item "Share a comprehensive rundown of the presented video, highlighting its main sequences."
            \item "Offer a thorough analysis of the video, discussing its various elements and storyline."
            \item "Explain the various aspects of the video before you, including the setting and actions."
            \item "Clarify the contents of the displayed video with great detail, focusing on its progression."
            \item "Characterize the video using a well-detailed description, capturing its essence and events."
            \item "Break down the elements of the video in a detailed manner, discussing its key components."
            \item "Walk through the important details of the video, describing its scenes and characters."
            \item "Portray the video with a rich, descriptive narrative, capturing its atmosphere and events."
            \item "Narrate the contents of the video with precision, focusing on its storyline and visuals."
            \item "Analyze the video in a comprehensive and detailed manner, discussing its themes and elements."
            \item "Illustrate the video through a descriptive explanation, painting a vivid picture of its content."
            \item "Examine the video closely and share its details, including the actions, characters, and setting."
            \item "Write an exhaustive depiction of the given video, capturing its essence and key moments."
        \end{itemize}
    \end{tcolorbox}
    \vspace{-2mm}
    \caption{\textbf{The list of instructions for detailed image and video description.}
    The image list is inherited from LLaVA \cite{llava}.
    The video list is generated by ChatGPT with examples of the image list.
    }
    \label{tab:detailed_describe_instructions}
\end{minipage}
\end{table*}

\end{document}